# Using Variable Threshold to Increase Capacity in a Feedback Neural Network

Praveen Kuruvada

Abstract: The article presents new results on the use of variable thresholds to increase the capacity of a feedback neural network. Non-binary networks are also considered in this analysis.

## Introduction

Information stored in a neural network is generally viewed as a pattern of activities in some models while in others, it is seen as information stored at specific location [1]-[4]. The processing abilities of a neuron are yet unknown, if the processing ability of a neuron is limited to process very low and fundamental information or if the neuron is capable of handling high level processing. A binary neural network has a capacity of 0.14N, where N being the number of neurons in the network. A binary neural network barely describes the complexity of a biological neural network, since a actual biological neuron not only carries electrical impulses but also they are associated with a train of voltage spikes. A non-binary neural network has a much higher capacity than that of a binary neural network but the computation complexity in dealing with these networks is high.

The questions of counter-intuitive and paradoxical aspects of memory storage [5]-[11] are not considered here. These questions relate to both quantum aspect of the physical law as well as the fact that memory storage implies selectivity.

Learning plays a major role in most of the neural network architectures. What do we mean when we actually say learning? Learning implies that a network or a processing unit is capable of changing its input/output behavior as a result of change in the environment. When the network is constructed the activation rule is usually fixed, since the input/output vectors cannot be changed, to change the input/output behavior the weights corresponding to the input vector has to be adjusted. Hence a learning algorithm is required for training the network to change the input/output behavior.

In this paper we discuss about a learning approach used to store memories in a network using a variable threshold approach. This approach determines the threshold corresponding to each neuron in a network. The Hebbian rule of learning is applied to neurons to store memories in feedback neural networks. The experimental results of using this approach over different size of networks are presented. This approach has been further applied to the B-Matrix approach of memory retrieval [12]. This is also of relevance in considering indexed memories [13]-[19]. The capacity and complexities of a non-binary neural network have been discussed [20],[21].

## Feedback networks

A model that assumes that the synaptic strength between any two neurons is changed based on the pre-synaptic neuron's persistent stimulation of the post-synaptic neuron was proposed by Donald Hebb in 1949. The Learning rule in the Hebbian model is given by

$$W_{(i,j)} = x_i . x_j'$$

Where x is the input vector and W is a weight n x n matrix containing the weights of successively firing neurons. A memory is considered to be stored in the network if

x=sgn(W.x)

Where sgn is the signum function. The signum function is sign(k) equals 1 when k is equal to or greater than zero and -1 for rest of the values. Such a network (figure 1) can have loops in the network i.e. signals can travel in both forward and backward directions. Feedback networks are powerful, dynamic (their state changes continuously until they reach a equilibrium point) and hence are complex networks. They remain at the equilibrium point until a change in the input occurs and a new equilibrium point is to be found.

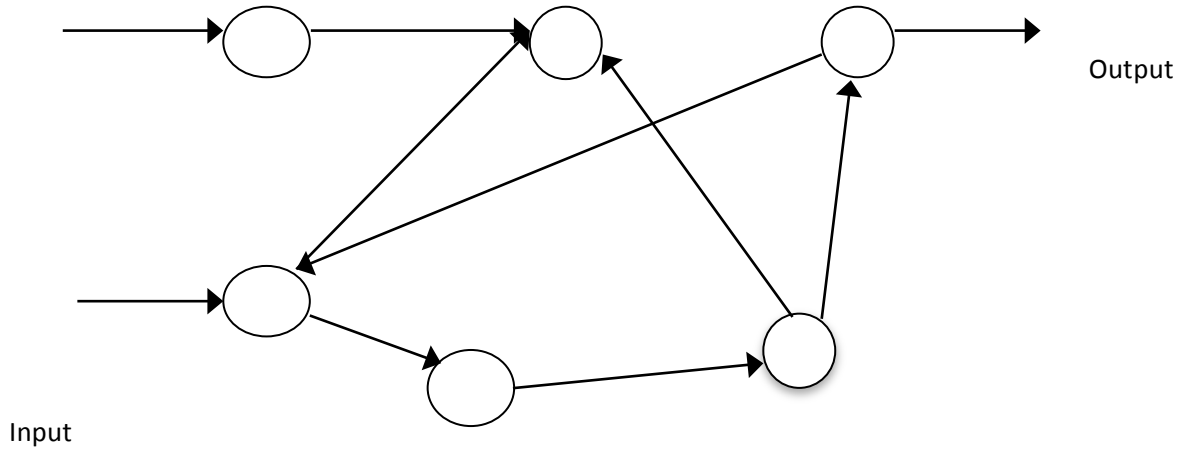

**Figure 1**. Feedback neural network

The Widrow-Hoff learning rule increases the memory storage capacity of the Hebbian network. In this rule the weights of the neurons that are stored in the network are adjusted to increase the possibility of retrieving those memories form the network. Initially the calculation of the error associated with the retrieval of memory form the network is done. Based on this error matrix obtained, the weights are adjusted such that the error associated with the particular memory is minimized. This process is continued until all the memories are stored in the network with minimal error or with a permissible threshold.

$W_{n+1} = W_n + \Delta(W_n)$ , where $\Delta(W_n) = \eta(x_i - W_n x_i)$, W is the weight matrix, $x_i$ is the present input, and $\eta$ is a small positive constant.

An error vector is estimated for each iteration of weighted matrix adjustment. Then an error term associated with these vectors is calculated and average this error term over the number of memories that are trained to the network. Defining a good error term is one of the major problems of this model.

**B-Matrix Approach**

The B-Matrix Approach [12], developed by Kak is a feedback network with indexed memory retrieval. The B-Matrix approach is a generator model for memory retrieval. In this the fragment length increases as the activity starts from one neuron and spreads to adjacent neurons. The fragment generated as a result is fed back to the network recursively until the memory is retrieved. The use of proximity matrix along with the B-Matrix approach was also shown by Kak [23].

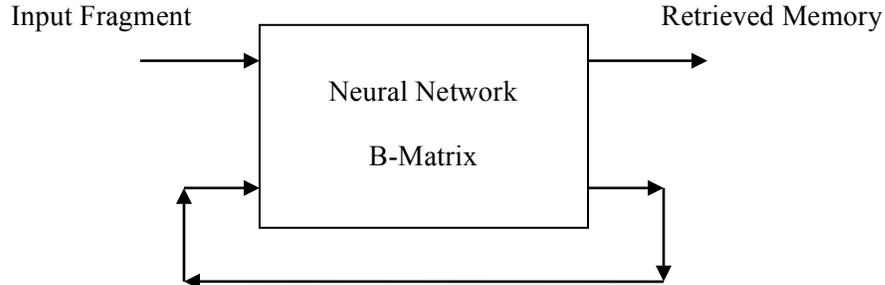

**Figure 2.** B-Matrix generator model

Recollection of memories in the B-Matrix Approach is by using the lower triangular matrix *B*, constructed as,

$T = B + B^t$

The proximity matrix stores the synaptic strength between the neurons. The activity starts from one neuron and spreads to additional neurons based on these synaptic strengths determined by the proximity matrix. Starting with the fragment $f^1$, the updating proceeds as:

$f^i = sgn(B \cdot f^{i-1})$,

Where $f^i$ is the $i^{th}$ iteration of the generator model. The $i^{th}$ iteration does not alter the i-1 iteration values but only produces the value of the $i^{th}$ binary index of the memory vector.

This model relies heavily on the geometric organization of the network. The proximity matrix provides this information of the geometric proximity of each neuron from every other neuron.

The B-Matrix could be viewed as a model that lies between feedback and feedforward neural networks since information in this model only flows in one direction. For quick training of feedforward neural networks using unary and similar coding, see [22]-[30].

The neural network of *n* neurons may be thought of as a three dimensional network of *n* nodes interconnected with each other with different synaptic strength between each node. We can construct a two dimensional graph of the network as a polygon of *n* sides with all diagonals connected and each corner being a node. For example, consider the neural network of 6 nodes as shown in Figure 3.

Let us assume without loss of generality that this network is already trained with a set of memories. When retrieving a memory from this network, the activity starts from one node and then spreads to the adjacent nodes as described by the proximity matrix. Assume that the activity starts at the second neuron and spreads from there on. If the synaptic order given by the proximity matrix is [2 5 3 1 4 6], then memory retrieval proceeds as shown in the figure.

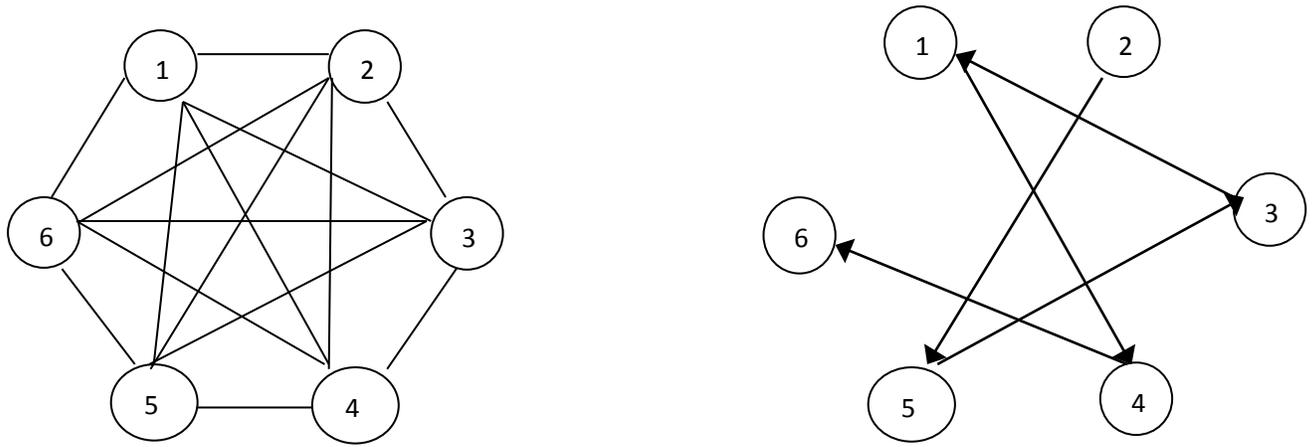

**Figure 3.** Graph showing the Neural Network and Activity Spread from Neuron 2.

Each neuron is capable of storing and retrieving a particular memory by the spread of activity as long as the network is not overwhelmed with information. As seen in the above example, it is possible to retrieve the right memories if we do know what the index of the neuron to be stimulated is, and what should be used to stimulate that selected neuron. Hence indexing plays a major role in the retrieval of memories. To better understand how a complex neural network might function, we introduce the concept of a sub-neural net and an indexing neural net. A different approach to indexing is provided in [13] , [17],[29] .

This model eliminates the need for a complete vector of memory needed for verification and a partial fragment of memory would be sufficient for recall. To illustrate this with an example, consider a person listening to a song. The next time he hears the song, he need not listen to the whole song to remember that this was the song he heard. All he needs is a small fragment of the song to help him recollect and maybe even sing the song.

Actual neural networks of the brain are far more complex than the models discussed in the literature and the processing abilities of each neuron is yet unknown. The basic neural network has a neural weight, which represents the strength of the connection for each synapse and a nonlinear threshold in the neuron soma (cell body). As the sum of the weighted inputs to the neuron soma increases there is relatively little response from the neuron until a critical threshold is reached by the neuron, at this point the neuron rapidly increases the output of its axon and fires. Hence different neurons have different threshold.

Here we take into consideration that neurons fire at different thresholds. Varying threshold values are learnt by the neurons as part of the training process. The advantages of doing so are shown both for the standard Hopfield model as well as for the B-matrix model.

**Variable Threshold Learning Approach**

In the variable threshold learning algorithm approach each neuron is assumed to have a different threshold, which, as we have argued, is likely to be more realistic than the uniform threshold neuron model. The threshold associated with each neuron is calculated in an iterative manner by increasing the threshold by 0.1 till the maximum numbers of neurons are stored in the network.

1) Identify the number of neurons in the network and the number of memories to be stored.

2) Find the T-Matrix associated with the memory vectors considered. The interconnection T-Matrix is constructed by multiplying the memory vectors with their transpose.
$$T = \Sigma \, x^{(i)} x^{(i)t}$$
Where the memories are binary column vectors ($x^i$), composed of $\{-1, 1\}$ and the diagonal terms are taken to be zero.

3) Find the threshold corresponding to each neuron in the network. This is done in an iterative manner where the threshold at each neuron is increase by a factor of 0.1 and the number of memories that could be stored is determined. This process is continued till the thresholds at each neuron is determined with which the maximum number of memories could be store into the network.

4) In the variable threshold approach a memory is considered to be stored into the network when
$$X^i = T_i(T \cdot x^{(i)}).$$

5) When new memories are brought into the network the thresholds are to be adjusted such that maximum memories can be stored into the network (repeat step 3).

The variable threshold approach shows a increase in the number of memories store in the network when a smaller network is considered. The model performs better than the fixed threshold method when a network of 1000 neurons is considered. The percentage of memories stored decreases when a very larger network is considered but at least stores as many memories as the fixed threshold approach. One of the main issues in this model is the overhead in calculating the threshold iteratively when new memories are considered into the network.

Consider the example of a network with 7 neurons and 5 memories are to be stored into the network.

Let's assume that the actual memories are the vectors given below:

$X^1$ = 1  1  1  1  -1  1  -1

$X^2$ = 1  -1  -1  1  -1  1  -1

$X^3$ = 1  -1  1  -1  1  -1  1

$X^4$ = -1  1  1  -1  -1  -1  1

$X^5$ = 1  -1  1  1  1  -1  1

The interconnection matrix or the T-matrix is calculated as

$T = \Sigma \, x^{(i)} x^{(i)t}$

$$\begin{pmatrix} 0 & -3 & 1 & 3 & 1 & 1 & -1 \\ -3 & 0 & 1 & -1 & -3 & 1 & -1 \\ 1 & 1 & 0 & -1 & 1 & -3 & 3 \\ 3 & -1 & -1 & 0 & -1 & 3 & -3 \\ 1 & -3 & 1 & -1 & 0 & -3 & 3 \\ 1 & 1 & -3 & 3 & -3 & 0 & -5 \\ -1 & -1 & 3 & -3 & 3 & -5 & 0 \end{pmatrix}$$

We can observe that when the memory vectors are multiplied with the T-Matrix approach using the fixed threshold approach only one memory could be stored in the network.

The memory stored in the network is

$X^3$= 1   -1  1 -1  1 -1  1

Apply the variable threshold approach to the above network would require to determine the thresholds corresponding to each neuron through an iterative process. The threshold corresponding to each neuron for the above example are found to be:

-7.9 is the Threshold at Neuron1

0.1 is the Threshold at Neuron2

-7.9 is the Threshold at Neuron3

-3.9 is the Threshold at Neuron4

4.1 is the Threshold at Neuron5

-7.9 is the Threshold at Neuron6

-9.9 is the Threshold at Neuron7

Applying these threshold values at each neuron and storing memories in the network would result in storing following memories

| | | | | | | | |
|---|---|---|---|---|---|---|---|
| $X^1$= | 1 | 1 | 1 | 1 | -1 | 1 | -1 |
| $X^2$= | 1 | -1 | -1 | 1 | -1 | 1 | -1 |
| $X^3$= | 1 | -1 | 1 | -1 | 1 | -1 | 1 |
| $X^4$= | -1 | 1 | 1 | -1 | -1 | -1 | 1 |

This method gives the same performance when the threshold is increased by a factor of 0.2. Up to 0.5, this change in the output is due to the sudden increase in the threshold values. This approach is further extended to the B-matrix approach of memory retrieval. In both these cases a Widrow-Hoff technique is used to learn the variable thresholds. The improvement is capacity is substantial.

**Extending to the B-Matrix Approach**

Recollection of memories in the B-Matrix Approach is by using the lower triangular matrix $B$, constructed as,

$T = B + B^t$

The proximity matrix stores the synaptic strength between the neurons. The activity starts from one neuron and spreads to additional neurons based on these synaptic strengths determined by the proximity matrix. Starting with the fragment $f^1$, the updating proceeds as:

$f^i = \text{sgn}(B \cdot f^{i-1})$,

Where $f^i$ is the $i^{th}$ iteration of the generator model. The $i^{th}$ iteration does not alter the i-1 iteration values but only produces the value of the $i^{th}$ binary index of the memory vector.

This model relies heavily on the geometric organization of the network .The proximity matrix provides this information of the geometric proximity of each neuron from every other neuron.

Applying the variable threshold approach to the B-matrix requires the use of thresholds corresponding to each neuron during the memory retrieval process (i.e. when the fragment vector is applied on the B-Matrix, the threshold corresponding to specific neuron is considered). Consider the above example used for storing memories The B-Matrix for the above network is constructed as $T = B + B^t$

$$\begin{pmatrix} 0 & 0 & 0 & 0 & 0 & 0 & 0 \\ -3 & 0 & 0 & 0 & 0 & 0 & 0 \\ 1 & 1 & 0 & 0 & 0 & 0 & 0 \\ 3 & -1 & -1 & 0 & 0 & 0 & 0 \\ 1 & -3 & 1 & -1 & 0 & 0 & 0 \\ 1 & 1 & -3 & 3 & -3 & 0 & 0 \\ -1 & -1 & 3 & -3 & 3 & -5 & 0 \end{pmatrix}$$

To retrieve memories from the network using the B-Matrix we consider a fragment of memory and apply it repeatedly to the network till the entire memory is retrieved.

$$\begin{pmatrix} 0 & 0 & 0 & 0 & 0 & 0 & 0 \\ -3 & 0 & 0 & 0 & 0 & 0 & 0 \\ 1 & 1 & 0 & 0 & 0 & 0 & 0 \\ 3 & -1 & -1 & 0 & 0 & 0 & 0 \\ 1 & -3 & 1 & -1 & 0 & 0 & 0 \\ 1 & 1 & -3 & 3 & -3 & 0 & 0 \\ -1 & -1 & 3 & -3 & 3 & -5 & 0 \end{pmatrix} \begin{bmatrix} -1 \\ \\ \\ \\ \\ \\ \end{bmatrix} = \begin{bmatrix} -1 \\ 1 \\ 1 \\ -1 \\ -1 \\ -1 \\ 1_7 \end{bmatrix}$$

$$\begin{pmatrix} 0 & 0 & 0 & 0 & 0 & 0 & 0 \\ -3 & 0 & 0 & 0 & 0 & 0 & 0 \\ 1 & 1 & 0 & 0 & 0 & 0 & 0 \\ 3 & -1 & -1 & 0 & 0 & 0 & 0 \\ 1 & -3 & 1 & -1 & 0 & 0 & 0 \\ 1 & 1 & -3 & 3 & -3 & 0 & 0 \\ -1 & -1 & 3 & -3 & 3 & -5 & 0 \end{pmatrix} \begin{bmatrix} 1 \\ 1 \end{bmatrix} = \begin{bmatrix} 1 \\ 1 \\ 1 \\ 1 \\ -1 \\ 1 \\ -1 \end{bmatrix}$$

$$\begin{pmatrix} 0 & 0 & 0 & 0 & 0 & 0 & 0 \\ -3 & 0 & 0 & 0 & 0 & 0 & 0 \\ 1 & 1 & 0 & 0 & 0 & 0 & 0 \\ 3 & -1 & -1 & 0 & 0 & 0 & 0 \\ 1 & -3 & 1 & -1 & 0 & 0 & 0 \\ 1 & 1 & -3 & 3 & -3 & 0 & 0 \\ -1 & -1 & 3 & -3 & 3 & -5 & 0 \end{pmatrix} \begin{bmatrix} 1 \\ -1 \\ -1 \end{bmatrix} = \begin{bmatrix} 1 \\ -1 \\ -1 \\ 1 \\ -1 \\ 1 \\ -1 \end{bmatrix}$$

$$\begin{pmatrix} 0 & 0 & 0 & 0 & 0 & 0 & 0 \\ -3 & 0 & 0 & 0 & 0 & 0 & 0 \\ 1 & 1 & 0 & 0 & 0 & 0 & 0 \\ 3 & -1 & -1 & 0 & 0 & 0 & 0 \\ 1 & -3 & 1 & -1 & 0 & 0 & 0 \\ 1 & 1 & -3 & 3 & -3 & 0 & 0 \\ -1 & -1 & 3 & -3 & 3 & -5 & 0 \end{pmatrix} \begin{bmatrix} 1 \\ -1 \\ 1 \\ -1 \end{bmatrix} = \begin{bmatrix} 1 \\ -1 \\ 1 \\ -1 \\ 1 \\ -1 \\ 1 \end{bmatrix}$$

For the example considered all the memories stored in the network can be retrieved at using the variable threshold approach. The best-case complexity for this approach is when all the memory vectors could be stored without any weights adjusted and worst case when weights need to be adjusted for each $X_{ij}$ memory vector.

**Quaternary/ Non-Binary Neural Network:**

In all the topics discussed above we have used a simplified assumption of a neural network i.e. a binary neural network. It is not easy to compare these binary neural networks to the neural networks in biological organisms, since the biological neurons not only carry electrical impulses of varying magnitude, but they are also associated with a train of voltage spikes. For example most images are not binary and conversion of these images to binary would cause a loss of information. Non-binary or *n-ary* neural networks were proposed [21] motivated by the complexity of biological neural networks. The next state of the neuron is determined by equations like

$$x_i = \sum_j T_{i,j} V_j$$

Consider the construction of a quaternary neural network [21] instead of a binary neural network. Such a network implements the same principles, as does a binary one, with the exception that the neurons now, map to a larger set of activations.

$$V_i = \begin{cases} -4 & x_i < -t \\ -1 & x_i < 0 \\ 1 & x_i < t \\ 4 & x_i > t \end{cases}$$

Here t, 0 and –t are the thresholds. To store maximum patterns in the network a learning algorithm similar to the delta rule can be applied to the network to modify the synaptic weights.

$$\Delta T_{ij} = c\ (V^s_i - V_i)\ V^s_j$$

Where $V_i$ is the output of neuron I when pattern $V^s$ is applied to the network and c is the learning constant if $V^s_i = V_i$ then none of the weights in the $i^{th}$ row need to be adjusted. Repeated application of this procedure will lead to a convergence of $V_i$ to $V^s_i$ provided that the convergence ratio t/c is large enough. If convergence ratio is large the equation may have to be applied many number of times to learn the pattern and if the convergence ratio is too small than it may not be even possible to learn the pattern.

**Storage capacity of a Neural Network:**

The storage capacity of a binary neural network according to the Hopfield network is 0.14N where N is the number of neurons considered. The storage capacity of a non-binary neural network is much higher than that of a binary neural network and so is the complexity associated in dealing with a non-binary neural network.

In a quaternary binary neural network the selection of the threshold value plays a vital role in determining the storage capacity. For any n-ary neural network determining the convergence ration t/c is given by

$$t/c = V_{max}^2 \, Vdiff(N-1)$$

Where $V_{max}$ is the maximum allowable magnitude for the output of a neuron and $Vdiff$ is maximum difference between any two "adjacent" output values.

For the quaternary model this formula gives the minimum value of t/c that will guarantee that any pattern can be stored.

Calculating the threshold 't' with the above for network with inputs 2, -2, 1, -1 would be 48. Experiments have been performed on 7 neurons. Each time a different pattern is considered. The value of learning constant c is kept 1 for all the experiments performed. The value of t has been varied over a large range. In the first test performed the number of neurons was set to 7. For each value of t 100 attempts have been made to store 1 to six random patterns. Similar test was also performed taking 9 neurons into consideration. The results of the test are tabulated below.

**Table 1:** Results of non-binary network capacity with 9 neurons

| t/c | Patterns | | | | | |
|---|---|---|---|---|---|---|
| | 1 | 2 | 3 | 4 | 5 | 6 |
| 96 | 97 | 93 | 93 | 88 | 82 | 57 |
| 144 | 100 | 94 | 97 | 93 | 69 | 46 |
| 192 | 100 | 97 | 97 | 87 | 69 | 31 |
| 240 | 100 | 100 | 96 | 83 | 53 | 25 |
| 288 | 100 | 100 | 89 | 77 | 54 | 13 |
| 336 | 100 | 98 | 81 | 58 | 36 | 11 |

Table 1: Results of non-binary network capacity with 7 neurons

| t/c | Patterns | | | | | |
|---|---|---|---|---|---|---|
| | 1 | 2 | 3 | 4 | 5 | 6 |
| 192 | 88 | 45 | 21 | 11 | 2 | 0 |
| 240 | 93 | 86 | 82 | 77 | 75 | 57 |
| 288 | 97 | 93 | 83 | 78 | 68 | 33 |
| 336 | 91 | 92 | 83 | 81 | 53 | 7 |
| 384 | 100 | 98 | 89 | 63 | 49 | 11 |
| 432 | 100 | 92 | 87 | 69 | 29 | 9 |

In the first experiment about 53% we were successful in storing patterns and were around 50% successful when nine neurons were considered. Form the above we can observe that for a given number of neurons the quaternary network could store more neurons than the binary model.

The use of the variable threshold approach was simulated in Java. The simulation takes in the number of neurons and memories to be considered and generates random numbers for the memory vector. From these randomly generated memory vectors we construct the T-Matrix.

Initially the simulation calculates the number of memories stored into the network using the fixed threshold approach. After finding the number of memories stored using the fixed threshold approach, the variable threshold approach is applied to the same set of memory vectors considered previously for the fixed threshold approach.

In finding the number of memories stored in the network using the variable threshold approach initially the threshold corresponding to each neuron in the network is calculated. Once the thresholds at each neuron is determined by a iterative approach the memories are stored in the network if

$$x = Ti(W.x)$$

where $T_i$ is the threshold at the $i^{th}$ neuron.

**Table 2:** Data for number of memories stored using fixed and variable threshold.

| Neurons | Memories | Fixed Threshold | Variable Threshold |
| --- | --- | --- | --- |
| 10 | 10 | 1 | 1 |
| 20 | 10 | 2 | 3 |
| 30 | 10 | 3 | 7 |
| 40 | 10 | 3 | 8 |
| 50 | 10 | 3 | 10 |
| 60 | 10 | 7 | 10 |
| 70 | 10 | 5 | 10 |
| 80 | 10 | 8 | 9 |
| 90 | 10 | 10 | 10 |
| 100 | 10 | 10 | 10 |

In the table above 10 memories are stored into the network with neurons from 10-100 are considered during each experiment. Each time the experiment is performed different set of memories are considered. The results of the case with 100 neurons is given below. The results scale up well from the n=10 case.

**Table 3:** Data for number of memories stored using fixed and variable threshold for n= 100

| Neuron | Memory | Fixed Threshold | Variable Threshold |
| --- | --- | --- | --- |
| 400 | 100 | 0 | 0 |
| 450 | 100 | 0 | 1 |
| 500 | 100 | 1 | 8 |
| 550 | 100 | 3 | 18 |
| 600 | 100 | 2 | 30 |
| 650 | 100 | 3 | 31 |
| 700 | 100 | 16 | 52 |
| 750 | 100 | 9 | 58 |
| 800 | 100 | 25 | 68 |
| 850 | 100 | 27 | 78 |
| 900 | 100 | 47 | 88 |
| 925 | 100 | 43 | 93 |
| 975 | 100 | 43 | 94 |
| 1000 | 100 | 61 | 95 |

From the graph above we can clearly observe that when neurons from 10- 100 are considered in storing tem memories, the use of a variable threshold approach could store more memories into the network.

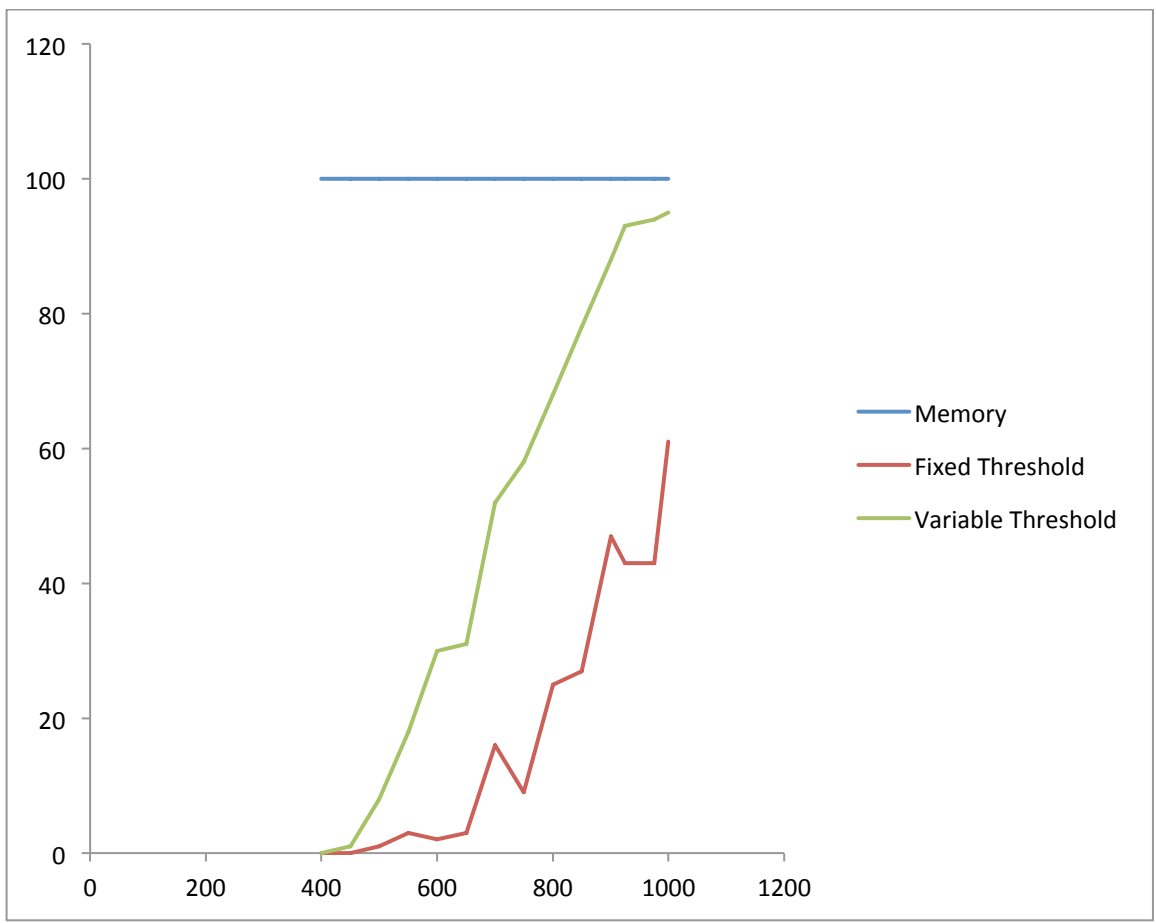

**Figure 4:** Fixed Threshold vs. Variable Threshold memory storage in a Neural Network when 100 memories are considered.

From all the experiments performed with 10 neurons to 1000 neurons we can observe that, from smaller networks to larger networks, when variable threshold algorithm results in storing a larger percentage of memories into the network when compared to fixed threshold approach.

**Applying Variable Threshold to B-Matrix Approach:**

Applying the variable threshold to the B-matrix requires use of thresholds corresponding to each neuron during the memory retrieval process (i.e. when the fragment vector is applied on the B-Matrix, the threshold corresponding to specific neuron is considered). This experiment has been performed several times considering various inputs and the results are displayed below in a table.

**Table 4:** Data for number of memories retrieved using fixed and variable threshold when 100 memories are considered.

| Neurons | Memories | Memories Stored | Fixed Threshold | Variable Threshold |
| --- | --- | --- | --- | --- |
| 10 | 10 | 1 | 1 | 1 |
| 20 | 10 | 4 | 2 | 3 |
| 30 | 10 | 7 | 3 | 6 |
| 40 | 10 | 8 | 3 | 8 |
| 50 | 10 | 10 | 3 | 8 |
| 60 | 10 | 10 | 7 | 8 |
| 70 | 10 | 10 | 5 | 7 |
| 80 | 10 | 9 | 8 | 9 |
| 90 | 10 | 10 | 9 | 10 |
| 100 | 10 | 10 | 10 | 10 |

From the graph above we can observe that when the variable threshold approach is applied to the B-Matrix approach of memory retrieval there is an increase in the percentage of memories retrieved form the network.

CONCLUSIONS

In this paper work we initially disused the Hebbian Model of memory storage in a neural network and the Widrow-Hoff model that was a extension to the Hebbian model in order to store more memories into a network. We then discussed the use of the variable threshold-learning approach of the neural network, which helps in storing more memories into the network. This approach has been applied on a neural network with hundred memories and thousand neurons and the increase in percentage of memory storage has been noted to be approximately around 15% on an average for different runs of the experiment.

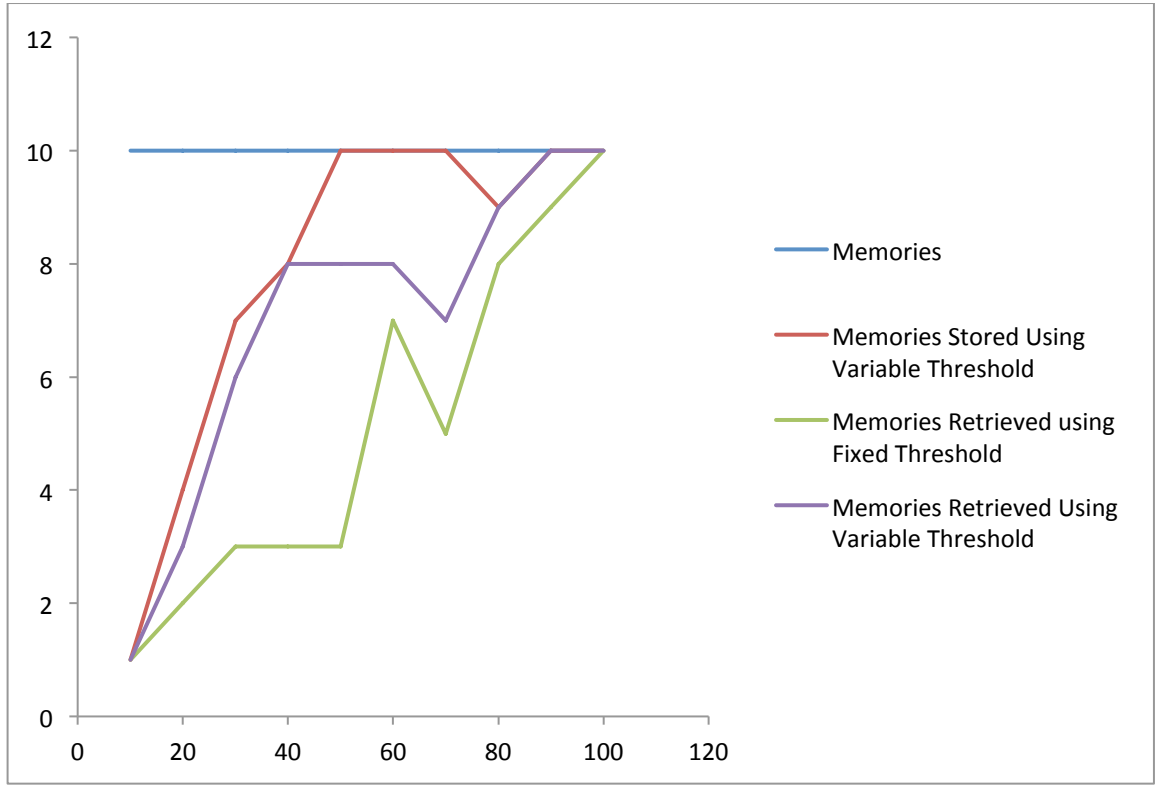

**Figure 5:** Graph representing the improvement in memory retrieval from a neural network when variable threshold is used.

This method has been further extended to the B-Matrix approach of memory retrieval. Using the variable threshold approach. Appling this algorithm related in an improvement in the overall percentage of memories retrieved from the network.

Future work involves further experiments on the behavior of this approach on a larger network or a non-binary neural network and finding efficient ways of defining thresholds at each point in the network in a more efficient manner in larger networks.

This work provides an idea of how the neural network performs when the threshold at each neuron varies form other in the network. Some questions that come to mind are, how efficient is this algorithm going to be when a larger network is considered or when a non-binary neural network is considered? And how much would be the effect on the thresholds at each neuron when new memories are brought into the network.